\documentclass[a4paper,twoside]{article}

\usepackage{epsfig}
\usepackage{calc}
\usepackage{amssymb}
\usepackage{amstext}
\usepackage{amsmath}
\usepackage{amsthm}
\usepackage{multicol}
\usepackage{pslatex}
\usepackage{apalike}
\usepackage{graphicx}
\usepackage{subfig}
\usepackage{booktabs, makecell,dcolumn}
\makeatletter
\newcommand{\clearsubcaptcounter}{\setcounter{sub\@captype}{0}}
\makeatother
\usepackage{SCITEPRESS}     

\begin{document}

\title{Symmetric Skip Connection Wasserstein GAN for High-Resolution Facial Image Inpainting}

\author{\authorname{Jireh Jam\sup{1},
		Connah Kendrick\sup{1},
		Vincent Drouard\sup{2},
		Kevin Walker\sup{2},
		Gee-Sern Hsu\sup{3} and 
		Moi Hoon Yap\sup{1}}
	\affiliation{\sup{1}Manchester Metropolitan University, Manchester, U.K}
	\affiliation{\sup{3}National Taiwan University of Science\&Technology, Taipei, Taiwan}
	\affiliation{\sup{2}Image Metrics Ltd, Manchester, U.K}}

\abstract{The state-of-the-art facial image inpainting methods achieved promising results but face realism preservation remains a challenge. This is due to limitations such as; failures in preserving edges and blurry artefacts. To overcome these limitations, we propose a Symmetric Skip Connection Wasserstein Generative Adversarial Network (S-WGAN) for high-resolution facial image inpainting. The architecture is an encoder-decoder with convolutional blocks, linked by skip connections. The encoder is a feature extractor that captures data abstractions of an input image to learn an end-to-end mapping from an input (binary masked image) to the ground-truth. The decoder uses learned abstractions to reconstruct the image. With skip connections, S-WGAN transfers image details to the decoder. Additionally, we propose a Wasserstein-Perceptual loss function to preserve colour and maintain realism on a reconstructed image.  We evaluate our method and the state-of-the-art methods on CelebA-HQ dataset. Our results show S-WGAN produces sharper and more realistic images when visually compared with other methods. The quantitative measures show our proposed S-WGAN achieves the best Structure Similarity Index Measure (SSIM) of 0.94.}

\onecolumn \maketitle \normalsize \setcounter{footnote}{0} \vfill

\section{\uppercase{Introduction}}
\label{sec:introduction}

\noindent Historically, inpainting is an ancient technique that was performed by professional artists to restore damaged paintings in museums. These defects (scratches, cracks, dust and spots) were inpainted by hand to restore and maintain the image quality. The evolution of computers in the last century and its frequent daily use has encouraged inpainting to take a digital format  \cite{efros1999texture,bertalmio2000image,criminisi2004region,pathak2016context,liu2018image,yang2017high,yan2018shift} as an image restoration technique. Image inpainting aims to fill in missing pixels caused by a defect based on pixel similarity information \cite{bertalmio2000image}.

The state-of-the-art approaches are two categories: conventional and deep learning-based methods. Conventional methods  \cite{efros1999texture,criminisi2004region,barnes2009patchmatch,sun2005image} use image statistics of best-fitting pixels to fill in missing regions (defects). However, these approaches often fail to produce images with plausible visual semantics. With the evolution in research, deep learning-based methods  \cite{pathak2016context,liu2018image,iizuka2017globally,yu2018generative,yan2018shift,yang2017high} encode the semantic context of an image into feature space and fill in missing pixels on images by hallucinations \cite{yang2017high} through the use of generative neural network.		
\begin{figure}
	\centering
	\subfloat[]{\includegraphics[width=0.117\textwidth]{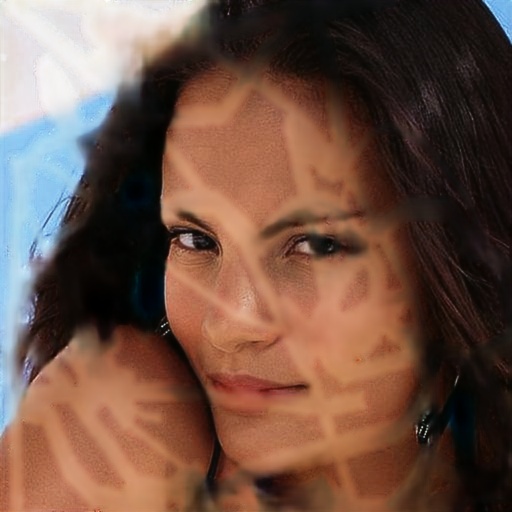}}
	\subfloat[]{\includegraphics[width=0.117\textwidth]{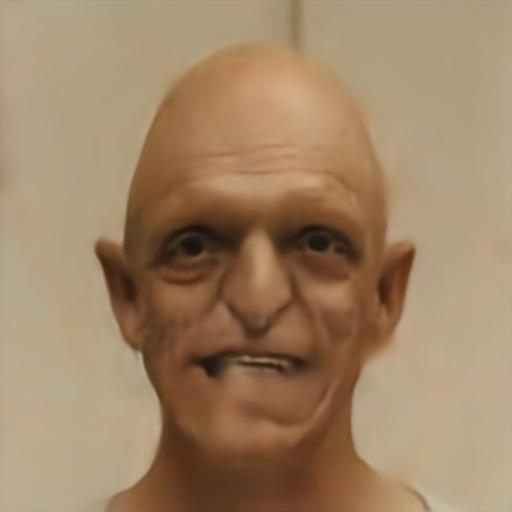}} 
	\subfloat[]{\includegraphics[width=0.117\textwidth]{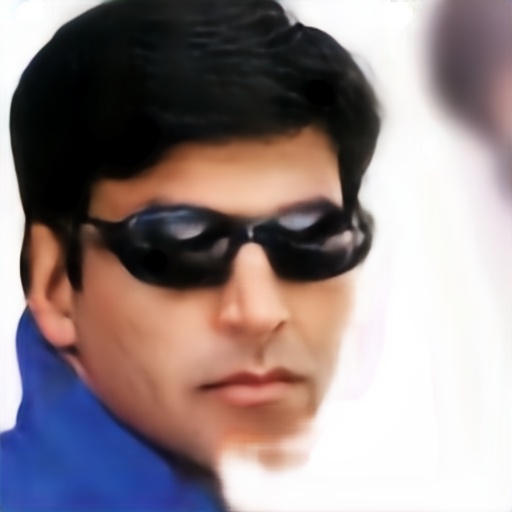}} 
	\subfloat[]{\includegraphics[width=0.117\textwidth]{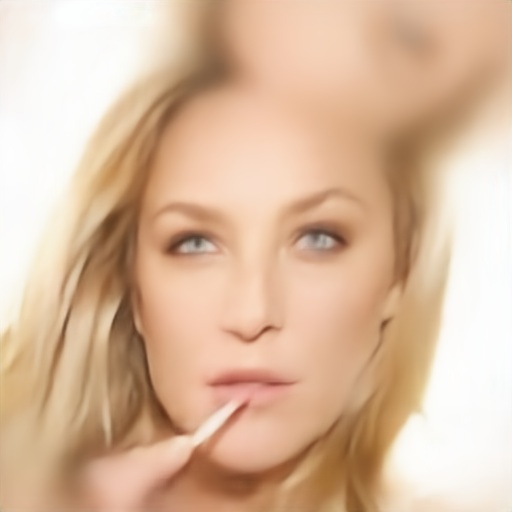}}
	\caption{\label{fig:limitations_Images} Images showing some issues by state of the art: (a) Poor performance on holes with arbitrary sizes; (b) Lack of edge-preserving technique; (c) Blurry artefacts; and (d) Poor performance on high-resolution images and image completion with mask at the border region.}	
\end{figure}
Although deep learning approaches achieve excellent performance in facial inpainting, there are some limitations of state of the art as illustrated in Figure~\ref{fig:limitations_Images}. These are cases, where Figure~\ref{fig:limitations_Images}(a) shows poor performance on holes with arbitrary sizes; Figure~\ref{fig:limitations_Images}(b) illustrates the lack of edge-preserving using the existing technique; Figure~\ref{fig:limitations_Images}(c) depicts the blurry artefacts; and Figure~\ref{fig:limitations_Images}(d) demonstrates the poor performance on high-resolution images and image completion with mask at border region. 

To correctly predict missing parts of a face and preserve its realism, we propose S-WGAN with the following contributions:
\begin{itemize}
	\item We propose a new framework with Wasserstein Generative Adversarial Network (WGAN) that uses symmetric skip connection to preserve image details.
	\item We define a new combined loss function based on RGB and feature space.
	\item  We demonstrate that our loss, combined with our S-WGAN, can achieve better results than the state-of-the-art algorithms. 
\end{itemize}	
\begin{figure*}[!h]
	\vspace{-0.2cm}
	\centering
	\includegraphics[width=15cm,height=8cm,keepaspectratio]{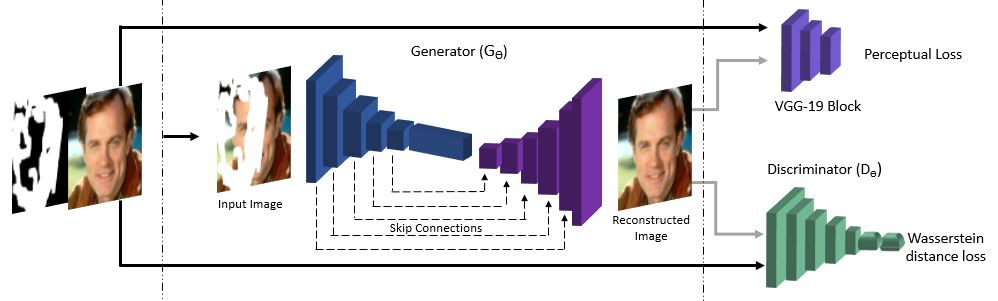}
	\caption{\label{fig:inpaint-framework} S-WGAN framework. The dilated convolution and deconvolution with the element-wise sum of feature maps (skip connection) combined with a Wasserstein network. The skip connections in the diagram ensure local pixel-level accuracy of the feature details to be retained.}
	\vspace{-0.1cm}
\end{figure*}
\section{\uppercase{Previous Work}}

\noindent Pathak et al. \cite{pathak2016context} proposed to use GANs \cite{goodfellow2014generative} with a context-encoder similar to \cite{vincent2010stacked,le2011optimization} and AlexNet  \cite{krizhevsky2012imagenet} for image inpainting despite poor hallucinations. Results show more artefacts and blur with randomised hole-to-image mask regions.
Iizuka et al. \cite{iizuka2017globally} used a local and global discriminator to assess coherency and consistency of predicted pixels, and replaced the fully-connected layer of the generator with dilated convolutions \cite{yu2015multi}.
Iizuka et al. \cite{iizuka2017globally} method failed to capture long-ranged textured information, however they used Poisson Blending by Perez et al. \cite{perez2003poisson} to process the output image. Yang et al. \cite{yang2017high} proposed a multi-scale neural patch synthesis based on style transfer \cite{johnson2016perceptual,ulyanov2016texture, li2016combining}, but failed to guarantee content and texture of high-resolution images with difficulty on irregular mask inpainting task.
Yeh et al. \cite{yeh2017semantic} introduced a spatial attention mechanism in deep convolutional GAN \cite{radford2015unsupervised} combined with context loss but this algorithm suffers misalignment on closest encoding in latent space. It performs poorly in handling of high-resolution and complex scene images. 
Li et al. \cite{li2017generative} used face parsing network combined with a generator (encoder-decoder) and two discriminators optimised by a semantic parsing loss to ensure local-global consistency and pixel fidelity. However, despite excellent performance, neighbouring pixels fail to establish spatial connections leading to colour inconsistencies.
Li et al. \cite{li2018learning} introduced reflection symmetry into face completion and used two networks, to establish a correspondence between missing pixels on two half-faces optimised using a symmetry loss defined on VGGFace \cite{parkhi2015deep}. However, this network fails to preserve structural information and is computationally costly. 

Liu et al. \cite{liu2018image} used partial convolution to replace typical convolutions \cite{ulyanov2018deep} with an automatic mask-updating step. This technique masks and renormalise convolutions to target only valid pixels. However, it performs poorly on sparsely structured images and binary masks with huge holes and no quantitative evaluation report on facial images. Yan et al. \cite{yan2018shift} used deep feature rearrangement by adding a particular shift-connection layer to the U-Net architecture \cite{ronneberger2015u}, but lacks efficiency with no guarantee in computational speed. Yu et al. \cite{yu2018generative} proposed a dual-stage network convolutional network combined with a contextual attention layer that learns the location of feature information from background patches to generate missing content. However, this network lacks pixel-wise consistency on high-resolution images. Liu et al. \cite{liu2019facial} proposed a multi-scale feature extraction powered by a multi-level generative network optimised by content and texture losses based on Mean Square Error ($\boldsymbol\ell_{2}$) and Structure Similarity Index (MS-SSIM), to capture features at various levels. This model struggles with larger masks and fails to preserve structure in unaligned facial images. Li et al. \cite{li2019face} proposed a nested GAN for facial inpainting, that uses a residual connection structure to transport information and interpolate feature map in deeper layer and shallow layer. Wang et al. \cite{wang2019laplacian} introduced a Laplacian approach based on residual learning \cite{he2016deep} to propagate high-frequency details and predict missing information. Despite the significant contributions by the methods above to the field of inpainting, the absence of preserved realism on facial images from a compact latent feature is still challenging due to larger and irregular masks. 

\section{\uppercase{Proposed Framework}}
Our proposed model uses skip connections with dilated convolution across the network, to perform image inpainting. We discuss the architecture and loss function of S-WGAN in the following sections
\subsection{Architecture}
\noindent 
\begin{figure*}
	\vspace{-0.2cm}
	\centering
	\clearsubcaptcounter
	\subfloat{\includegraphics[width=0.24\textwidth]{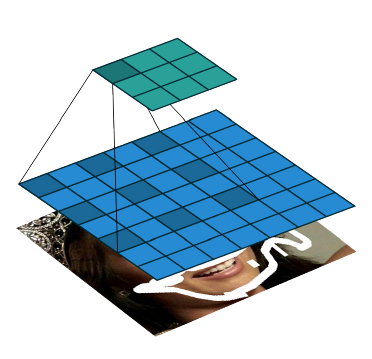}}
	\subfloat{\includegraphics[width=0.24\textwidth]{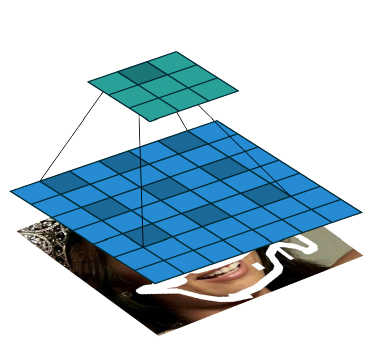}}
	\subfloat{\includegraphics[width=0.24\textwidth]{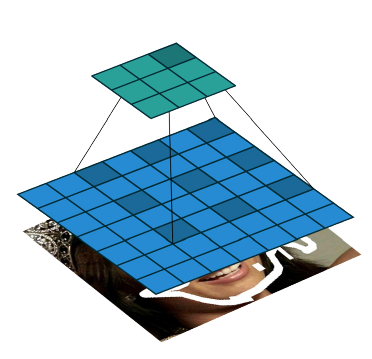}}
	\subfloat{\includegraphics[width=0.24\textwidth]{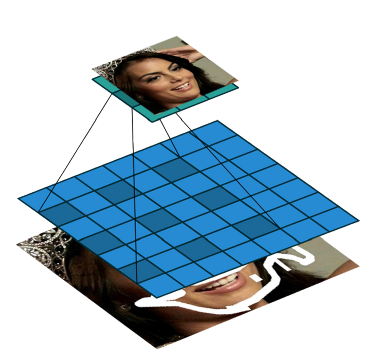}}
	\caption{\label{fig:dilation} Illustration of dilated convolution process. Convolving a $3 \times 3$ kernel over a $7 \times 7$ input with a dilation factor of 2 (i.e., $i = 7$, $k = 3$, $d_{r} = 2$, $s = 1$ and $p = 0$) \cite{dumoulin2016guide}.  The accretion of receptive field is in linearity with the parameters \cite{yu2015multi}. A $5\times5$ kernel will have the same receptive field view as over a $7 \times 7$ input at dilation rate=2 whilst only using 9 parameters over a $512 \times 512$ input.}
	\vspace{-0.1cm}
\end{figure*}
Figure~\ref{fig:inpaint-framework} shows the overall framework of our proposed S-WGAN. The network is designed to have a generator ($G_{\theta}$) and a discriminator ($D_{\theta}$). We define $G_{\theta}$ as an encoder-decoder framework with dilated convolutions and symmetric skip connections.  Figure~\ref{fig:dilation} shows the process of dilated convolution. Dilated convolutions \cite{yu2015multi}, combined with skip connections, are critical to the design of our model as:
\begin{itemize}
	\item It broadens the receptive fields to capture more contextual information without parameter accretion and computational complexity, which are preserved and transferred by skip connections to corresponding deconvolution layers.
	\item It detects fine details and maintains high-resolution feature maps, and achieves end-to-end feature learning with a better local minimum (high restoration performance). 
	\item It has shown considerable improvement of accuracy in segmentation task \cite{yu2015multi,chen2017deeplab,chen2017rethinking}.
\end{itemize}

\par{\textbf{Generator ($G_{\theta}$)}} The effectiveness of feature encoding is improved by having an encoder of ten-convolutional layers, with a kernel size of 5 and dilation rate of 2, designed to match the size of the output image. This technique enables our model to learn larger spatial filters and help reduce volume \cite{rosebrock_dl4cv}.
Each block of convolution in exception of the final layer has Leaky ReLU activation and max-pooling operation of pool size $2 \times 2$. We apply a dropout regularisation with a probability value of 0.25 in the 4th and final layer of the encoder. The dropout layer randomly disconnects nodes and adjust the weights to propagate information to the decoder without overfitting. 

\par{\textbf{Decoder}}
The decoder are five blocks of deconvolutional layers, with learnable upsampling layers that recover image details using the same kernel size and dilation rate of the generator. The corresponding feature maps in the decoder are asymmetrically linked by element-wise skip connections to reach an optimum size. The final layer in the decoder is Tanh activation.
\par{\textbf{Dilated Convolutions:}}
We express the dilated convolution based on the network input in Equation~\ref{eq:dilated}:
\begin{equation}
{\mathrm \mathcal{I}'_{(m,n)}}=\sum_{i=1}^{M}\sum_{j=1}^{N}(\mathbf{M}_{I})(m+d_{r}\times i, n+d_{r}\times j) \ast \omega_{(i,j)}   \label{eq:dilated}
\end{equation}

where $I'_{(m,n)}$ is the output feature map of the dilated convolution from the input $\mathbf{M}_{I}=(I \odot (1-M))+M$ and the filter is given by $\omega_{(i,j)}$. The dilation rate parameter ($d_{r}$) reverts to normal when $d_{r}=1$. 

It is advantageous to use dilated convolution compared to using typical convolutional layers combined with pooling. The reason for this is that a small kernel size of $\textit{k} \times \textit{k}$ can enlarge into $\textit{k} + (\textit{k}-1)(d_{r}-1)$ based on the dilated stride $d_{r}$, thus allowing a flexible receptive field of fine-detail contextual information while maintaining high-quality resolution.

The inpainting solver $G_{\theta}$ may result in predictions $G_{\theta}(\mathbf{\hat{z}}) $ of the missing region, that may be reasonable or ill-posed. We include as part of our network $D_{\theta}$, adopted from \cite{arjovsky2017wasserstein} to provide improved stability and enhanced discrimination for photo-realistic images. With ongoing adversarial training, the discriminator is unable to distinguish real data from fake ones. 
Equation~\ref{eq:completedImage} shows the reconstruction of the image during training from $G_{\theta}$:
\begin{equation}
G_{\theta}(I_{R}) = I \odot M +(1-M)\odot G_{\theta}(\mathbf{\hat{z}})  \label{eq:completedImage}
\end{equation}
where $I_{R}$ is the reconstructed image, $I$ is the ground-truth, $(\mathbf{\hat{z}}) $ is the predictions, $\odot$ is the element-wise multiplication and $M$ is the binary mask, represented in 0 and 1. In our case 0 is the context of the entire image and 1 is the missing regions. 

Equation~\ref{eq:wgan} adopted from \cite{arjovsky2017wasserstein} refers to the Wasserstein discriminator. 
\begin{equation}
\max_{D}V_{WGAN}=E_{x \sim p_{r}}[(D_{\theta}(I)]- \\ E_{z\sim p_{z}}[D(G_{\theta}(I_{R}))] 
\label{eq:wgan}
\end{equation}
where D is the discriminator and $P_{r}$ is real data distribution. G is the generator of our network and $P_{z}$ is the  distribution.

\subsection{Loss function}
\paragraph*{\textbf{Perceptual loss} }
Instead of using the typical  $\boldsymbol\ell_{2}$-loss function used in \cite{pathak2016context}, we use a new combination of loss functions, luminance ($L_{l}$) and feature loss. Pixel-wise reconstruction and feature space loss are not new to inpainting \cite{yeh2017semantic,yu2018generative, johnson2016perceptual}.
We define a luminance guided $L_{l}$ that uses $\boldsymbol\ell_{1}$-loss as a base to compute the loss using a range of constant pixel values in the RGB space. This preserves colour and luminance and does not over penalise large errors \cite{zhao2016loss}. We use the $L_{l}$ to adjust our perceptual loss, thus minimising any error $\textgreater$1. Also, the $L_{l}$  allows better evaluation of the predictions to match the ground-truth.
More specifically, we express the luminance loss ($L_{l}$) based on $\boldsymbol\ell_{1}$ as: 
\begin{equation}
\label{eq:msaloss}
L_{l} =|| K \odot (x_{i} - \mathbf{\hat{z}}_{i})||_1 
\end{equation}
where $i$ is the pixel index with $x_{i}$ and $\mathbf{\hat{z}}_{i}$ as pixel values of the ground-truth and the predictions, constraint by a constant K.
Our feature loss
$L_{f}$ is a feature based $\boldsymbol\ell_{2}$-loss, rather than being computed directly on the image we computed the loss in a feature space. To achieve this, we adopt a pre-trained VGG-16 model trained on ImageNet \cite{krizhevsky2012imagenet}, and use it  
as a feature extractor in our loss function. More specifically we use the output of block3-convolution3 of this model to generate image feature. We use the $\boldsymbol\ell_{2}$ as base to compute our loss function, which is the same as the perceptual loss proposed by Johnson et al. \cite{johnson2016perceptual}. 
The advantage of using feature space is that a particular filter determines the extraction of feature maps from low-level to high-level, sophisticated features. To reconstruct quality images, we compute our loss function with 
feature maps determined by block3-conv3, resized to the same size as masks and generated images. The reason is that using another output for example block4-conv4 or block5-conv5 will result in poor quality, as the network
starts to expand the view at these layers due to more filters used. 
Our feature loss is expressed as follows:
\begin{equation}
\begin{aligned}[b]
L_{f} =  {1\over N}\sum_{i \epsilon \phi} (\phi([{M_{I}}_{i}] -[{I_{R}}_{i}]))^2 
\label{eq:ploss-function}
\end{aligned}
\end{equation}
where $M_{I}$ is the input image,  $I_{R}$ is the reconstructed image and $N$ is dimensions obtained from $\phi$ feature maps with high-level representational abstractions extracted from the  third block convolution layer.
By combining $ L_{l}$ and $ L_{f}$ we obtained:
\begin{equation}
L_{p} = L_{l} + L_{f}  
\label{eq:loss-percep}
\end{equation} 
By using $L_{p}$ the model learns to produce finer details in the predicted features and output without any blurry artefacts. 
We add the Wasserstein loss ($L_{w}$) improves convergence in GANs and its the mean difference between two images.
Finally the entire model trains end-to-end with back-propagation and uses the global Wasserstein-perceptual loss function ($L_{wp}$) defined in Equation~\ref{eq:loss-function}, to optimise $G_{\theta}$ and $D_{\theta}$ to learn reasonable predictions.
Our goal is to reconstruct an image $I_{R}$ from $M_{I}$ by training the generator $G_{\theta}$ to learn and preserve image details.
\begin{equation}
L_{wp} = L_{w} + L_{p}  
\label{eq:loss-function}
\end{equation}
\begin{figure*}
	\vspace{-0.2cm}
	\centering
	\subfloat[]{\includegraphics[width=0.168\textwidth]{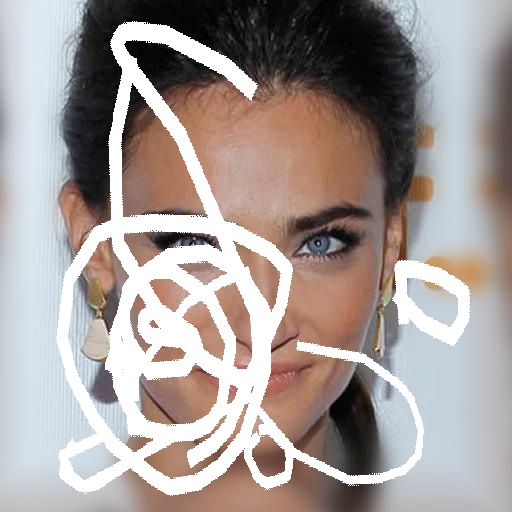}} 
	\subfloat[]{\includegraphics[width=0.168\textwidth]{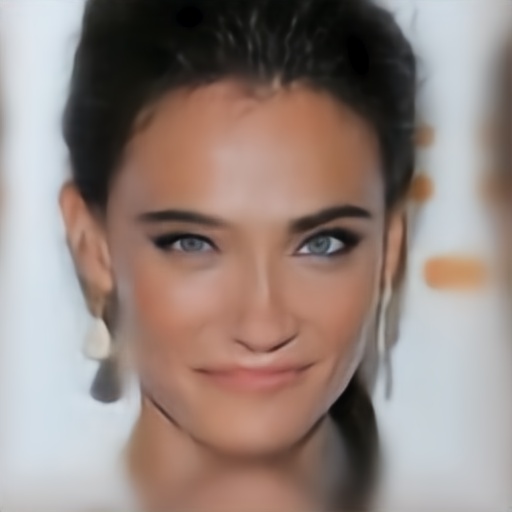}}
	\subfloat[]{\includegraphics[width=0.168\textwidth]{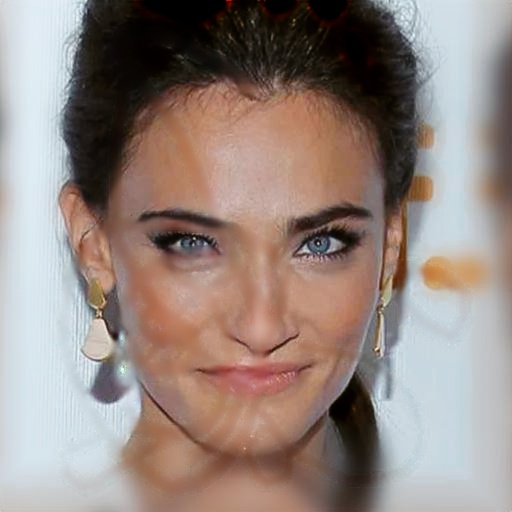}}
	\subfloat[]{\includegraphics[width=0.168\textwidth]{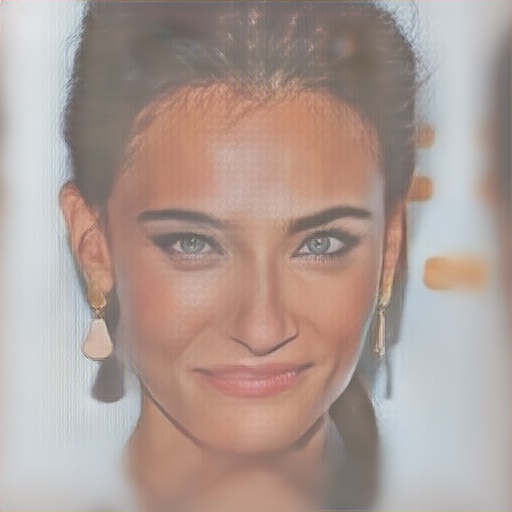}} 
	\subfloat[]{\includegraphics[width=0.168\textwidth]{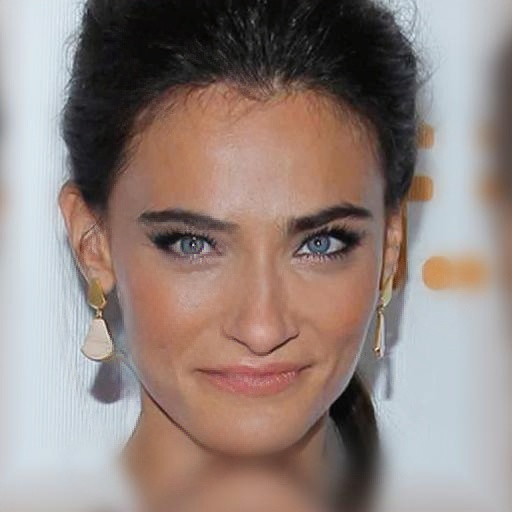}} 
	\subfloat[]{\includegraphics[width=0.168\textwidth]{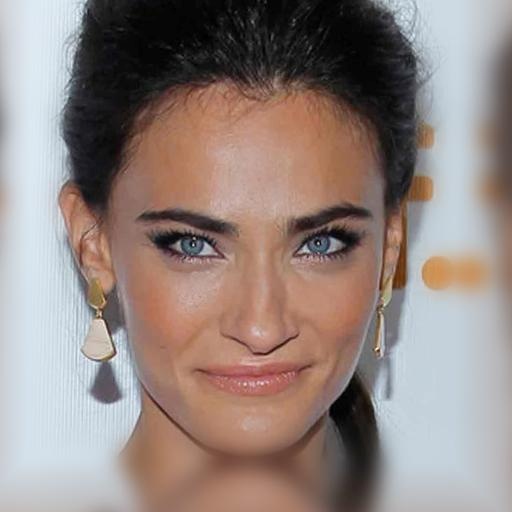}}\\[-5ex]
	\clearsubcaptcounter
	\subfloat[]{\includegraphics[width=0.168\textwidth]{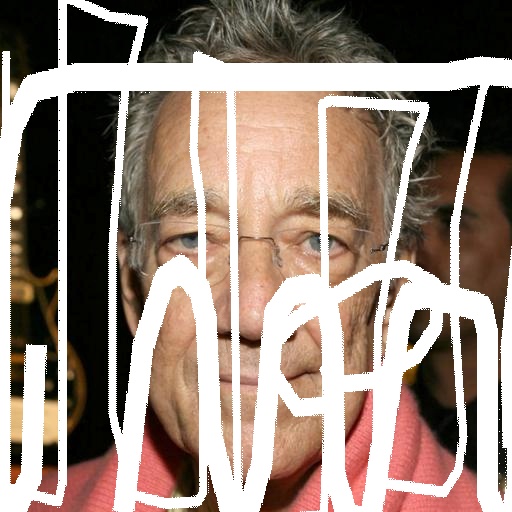}} 
	\subfloat[]{\includegraphics[width=0.168\textwidth]{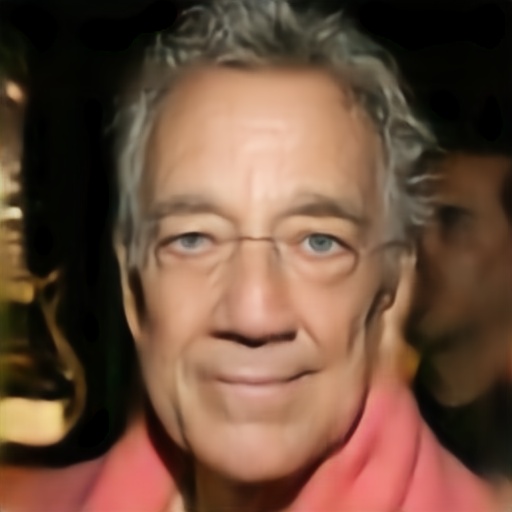}}	
	\subfloat[]{\includegraphics[width=0.168\textwidth]{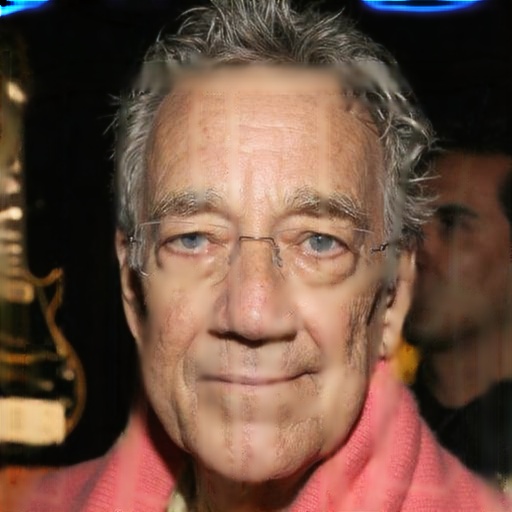}}
	\subfloat[]{\includegraphics[width=0.168\textwidth]{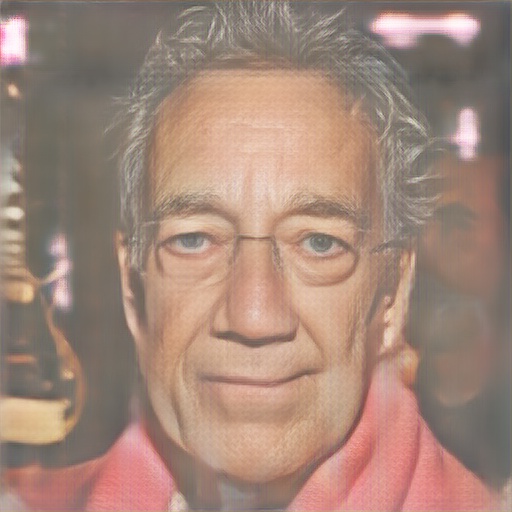}} 
	\subfloat[]{\includegraphics[width=0.168\textwidth]{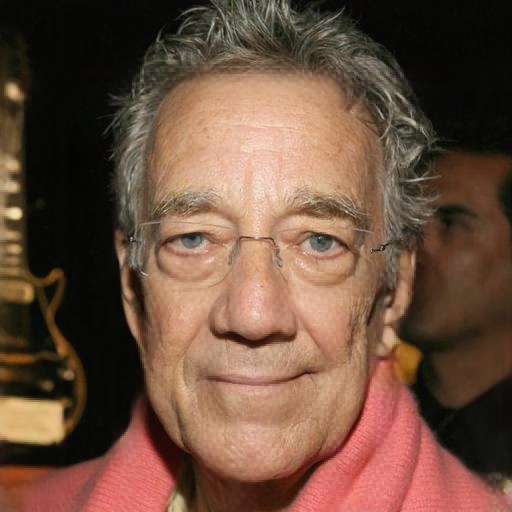}} 
	\subfloat[]{\includegraphics[width=0.168\textwidth]{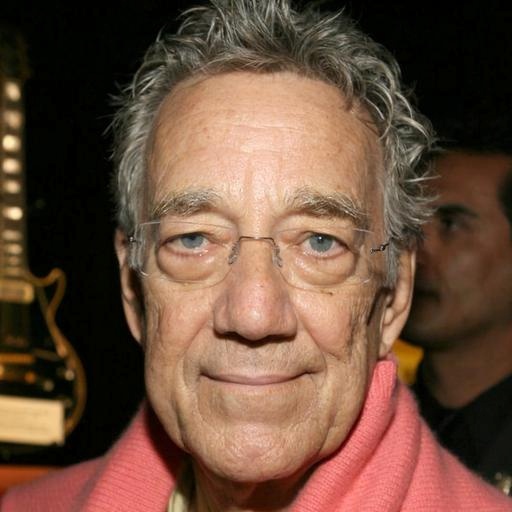}}\\[-5ex]
	\clearsubcaptcounter
	\subfloat[\textbf{INPUT}]{\includegraphics[width=0.168\textwidth]{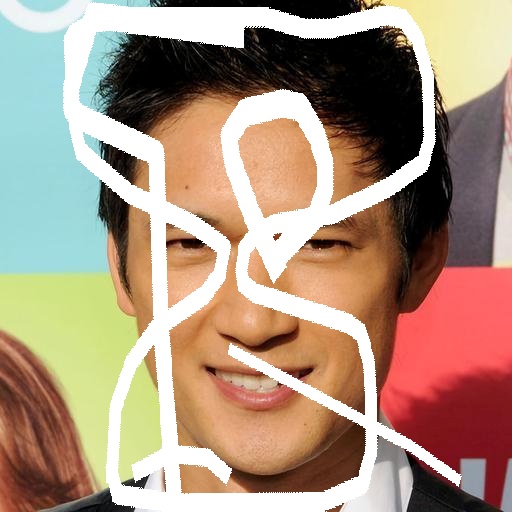}} 
	\subfloat[\textbf{CE}]{\includegraphics[width=0.168\textwidth]{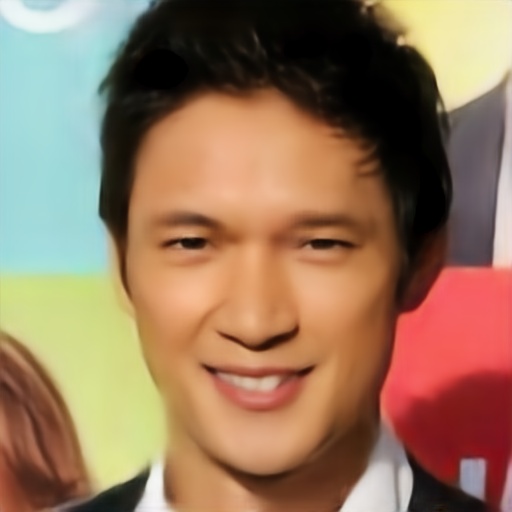}}	
	\subfloat[\textbf{PConv}]{\includegraphics[width=0.168\textwidth]{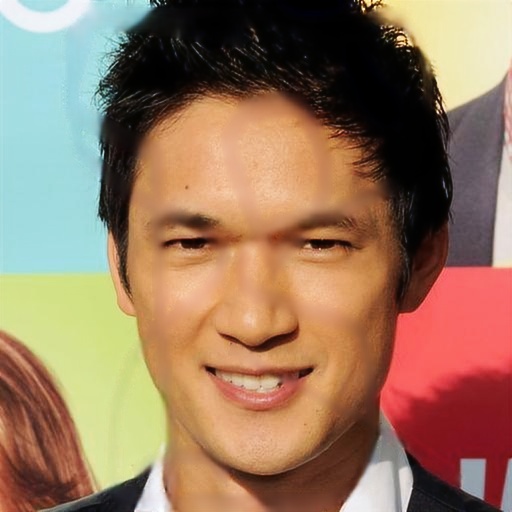}}
	\subfloat[\textbf{WGAN}]{\includegraphics[width=0.168\textwidth]{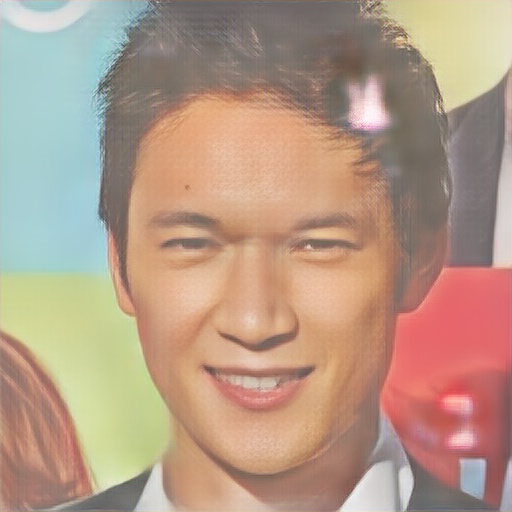}} 
	\subfloat[\textbf{S-WGAN}]{\includegraphics[width=0.168\textwidth]{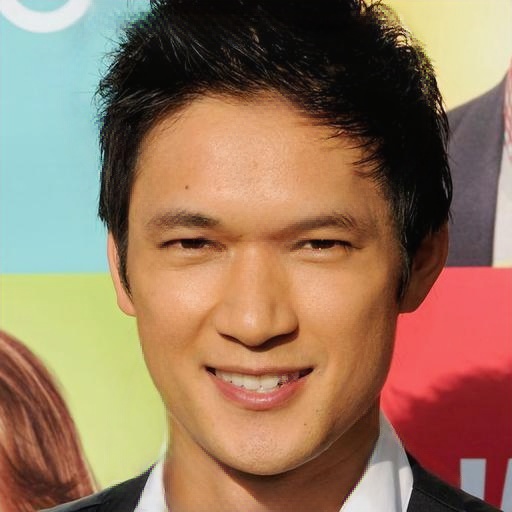}} 
	\subfloat[\textbf{GT}]{\includegraphics[width=0.168\textwidth]{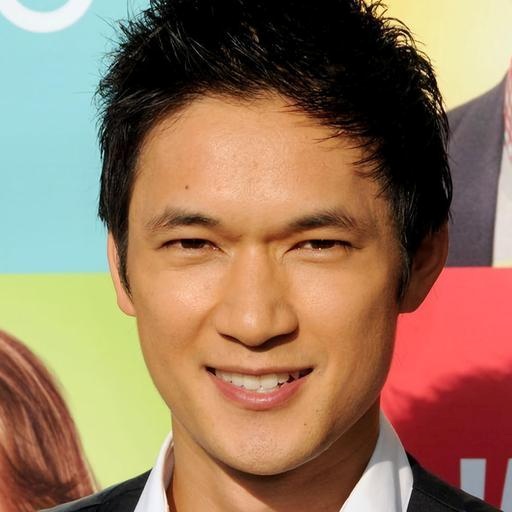}}		
	\caption{\label{fig:results}Qualitative comparison of our proposed \textbf{S-WGAN} with the state-of-the-art methods on CelebA-HQ: (a) \textbf{Input masked-image}; (b) \textbf{CE} \cite{pathak2016context}; (c) \textbf{PConv} \cite{liu2018image}; (d)  \textbf{WGAN}; (e)  \textbf{S-WGAN} (proposed method); and (f) Ground-truth image.}	
	\vspace{-0.5cm}
\end{figure*}
\section{\uppercase {Experiment}}
This section describes the dataset, binary masks and the implementation.
\subsection{Dataset and irregular binary mask}
\begin{figure}[h!]
	\centering
	\subfloat[]{\includegraphics[width=0.117\textwidth]{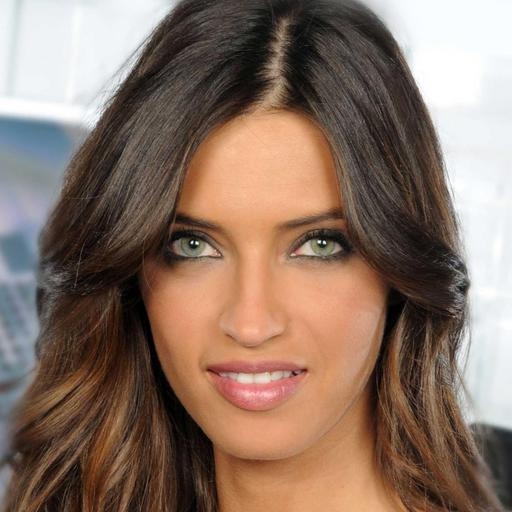}} 
	\subfloat[]{\includegraphics[width=0.117\textwidth]{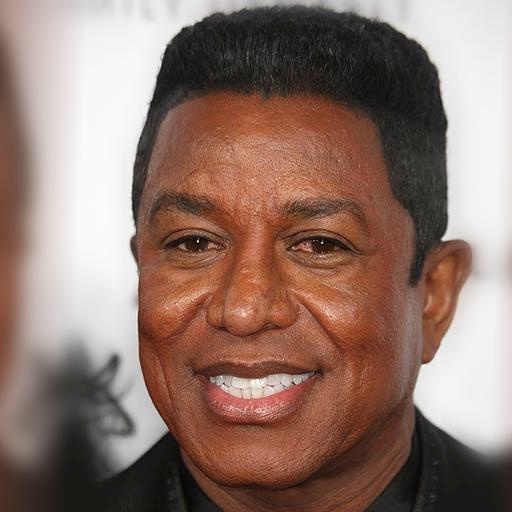}}
	\subfloat[]{\includegraphics[width=0.117\textwidth]{1822o.jpg}} 
	\subfloat[]{\includegraphics[width=0.117\textwidth]{24o.jpg}}	
	\caption{\label{fig:sampleImages}Sample images from CelebA-HQ Dataset \cite{karras2017progressive}.}	
\end{figure}

Our experiment focuses on high-resolution face images and irregular binary masks. The benchmark dataset for high-resolution face images is CelebA-HQ dataset  \cite{karras2017progressive}, which was curated from the CelebA dataset \cite{liu2018large} and contained 30,000 images. Figure~\ref{fig:sampleImages} shows a few samples from the CelebA-HQ dataset.

To create irregular holes on images, we use the Quickdraw irregular mask dataset  \cite{iskakov2018semi}, available for public use and is divided into 50,000 train and 10,000 test masks. The images are of size $512 \times 512$ pixels. 
\begin{figure}
	\centering
	\subfloat[\textbf{GT}]{\includegraphics[width=0.16\textwidth]{157o.jpg}} 
	\subfloat[\textbf{Mask}]{\includegraphics[width=0.16\textwidth]{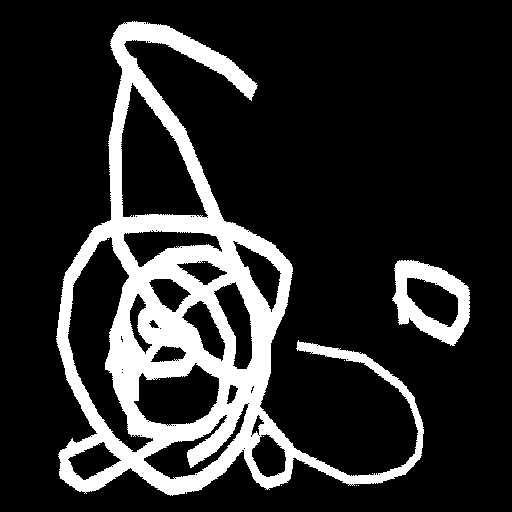}}
	\subfloat[\textbf{Masked-Image}]{\includegraphics[width=0.16\textwidth]{157m.jpg}} 
	\caption{\label{fig:matrixMul} Process of input generation: a) CelebA-HQ image; b) Binary mask image  \cite{iskakov2018semi}; and c) Corresponding masked image (input image).}	
\end{figure}

\subsection{Implementation}
We used the Keras library with TensorFlow backend to implement and design our network. With our choice of the dataset, we followed the experiment settings of state of the art \cite{liu2018image} and split our data into 27,000 images for training and 3,000 images for testing.

We perform normalised floating-point representation on the image to set the intensity values of the pixels in the range -1,1 and apply the mask on the image to obtain our input, as shown in Figure~\ref{fig:matrixMul}. 
We initialize pre-trained weights from VGG-16 to compute our loss function. We use a learning rate of $10^{-4}$ in $G_{\theta}$ and $10^{-12}$ in $D_{\theta}$ and optimise the training process using the Adam optimiser \cite{kingma2014adam}.
We use a Quadro P6000 GPU machine to train these models. According to our hardware conditions, we use a batch-size of 5 in each epoch for input images with shape $512 \times 512 \times 3$. It takes 0.193 seconds to predict missing pixels of any size created by binary mask on an image and ten days to train 100 epochs.
\begin{figure*}
	\vspace{-0.2cm}
	\centering
	\subfloat[\textbf{INPUT}]{\includegraphics[width=0.168\textwidth]{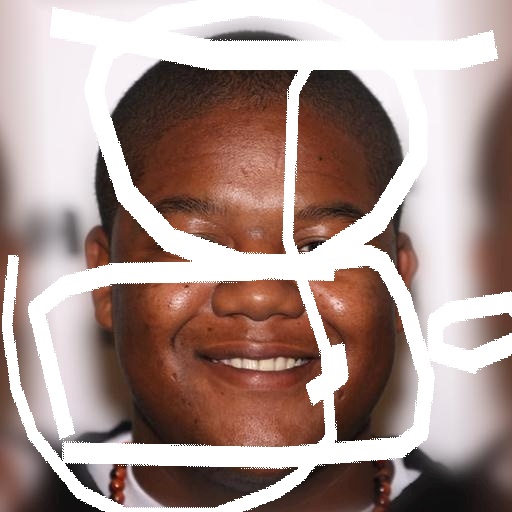}} 
	\subfloat[\textbf{WGAN}]{\includegraphics[width=0.168\textwidth]{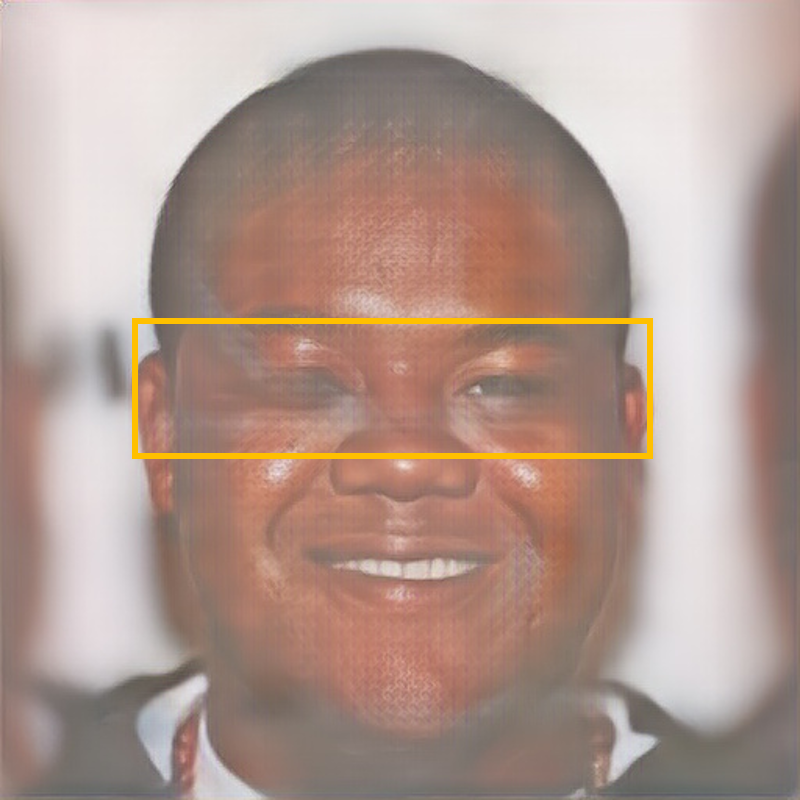}}
	\subfloat[\textbf{WGAN-S}]{\includegraphics[width=0.168\textwidth]{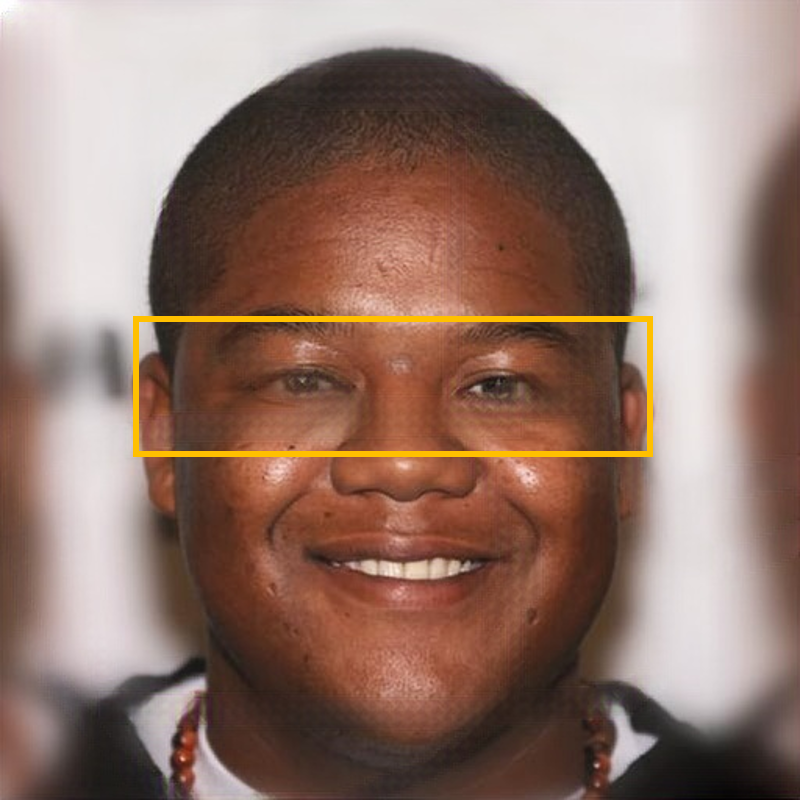}} 
	\subfloat[\textbf{WGANSD}]{\includegraphics[width=0.168\textwidth]{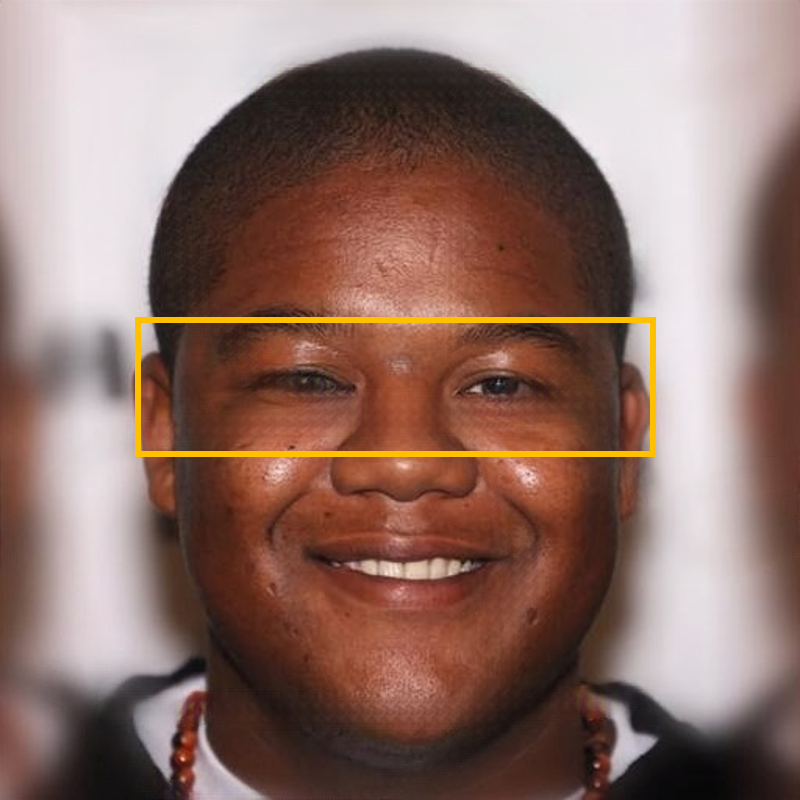}}
	\subfloat[\textbf{S-WGAN}]{\includegraphics[width=0.168\textwidth]{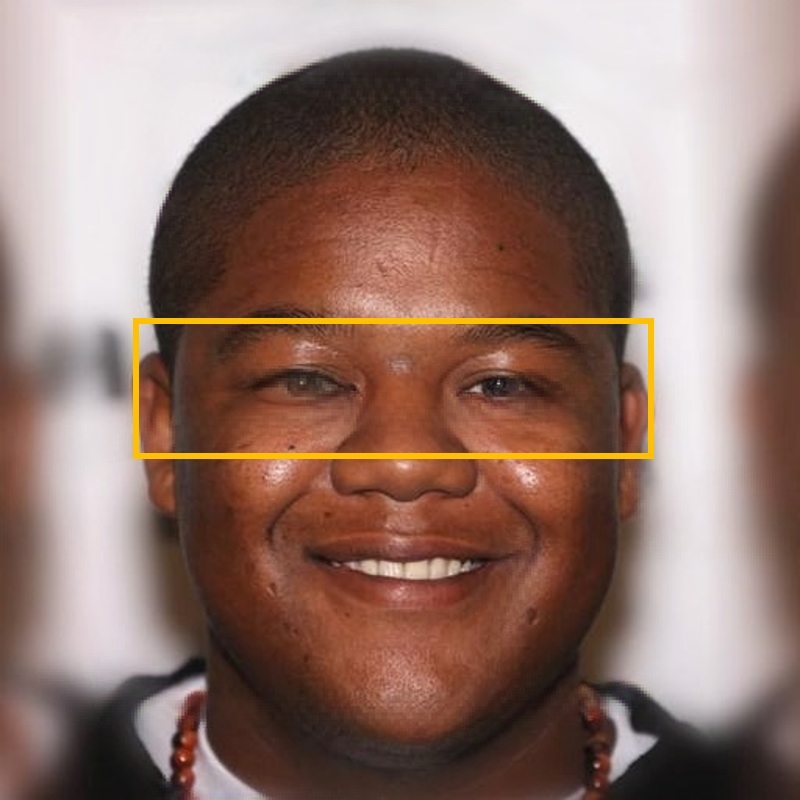}}
	\subfloat[\textbf{GT}]{\includegraphics[width=0.168\textwidth]{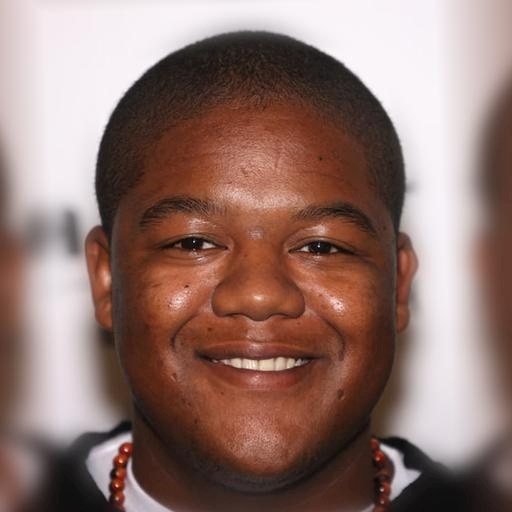}}
	\caption{\label{fig:ablation} Qualitative comparison of results using different architectures \cite{johnson2016perceptual} on CelebA-HQ \cite{karras2017progressive}. (a) Input masked image (b) Inpainted image by WGAN (c) Improved WGAN with skip connections (WGAN-S) (d) Improved WGAN with skip connection and dilated convolution (WGAN-SD) (e) Complete network with $L_{p}$ (f) Ground-Truth image. The yellow box indicates the region where other models failed to inpaint successfully completely. This region in (e) shows the effectiveness of $L_{p}$ on the inpainted image.}
	\vspace{-0.2cm}
\end{figure*}
\section{Results}
We assess the performance of the inpainting methods qualitatively and quantitatively in this section.
\subsection{Qualitative Comparisons}
Consider the importance of visual and semantic coherence; we conducted a qualitative comparison of our test dataset. First, we implemented a WGAN approach with $L_{f}$ and $L_{w}$. We observed an induced pattern and pitiable colour on the images, as shown in Figure~\ref{fig:results}(d). We introduced dilated convolution, skip connections combined with end-to-end training using $L_{wp}$ to handle the induced pattern and match the luminance of the original images. 

We compare our model with three popular methods:
\begin{itemize}
	\item \textbf{CE}: Context-Encoder method by Pathak et al. \cite{pathak2016context}. 
	\item \textbf{PConv}: Image Inpainting for irregular holes using partial convolutions by Liu et al. \cite{liu2018image}.
	\item \textbf{WGAN}: Wasserstein GAN method with perceptual loss.
\end{itemize}

We test our S-WGAN against state of the art on CelebA-HQ $512 \times 512 $ test dataset and show the results in Figure~\ref{fig:results}. Based on visual inspection, Figure~\ref{fig:results}(b) illustrates blurry generated by the Pathak et al.'s CE method \cite{pathak2016context}. On the other hand, PConv \cite{liu2018image} generates clear images but with residues of the binary mask left on the images as shown in Figure~\ref{fig:results}(c). WGAN induced pattern and low-contrast images, shown in Figure~\ref{fig:results}(d). Overall, our proposed S-WGAN, as shown in Figure~\ref{fig:results}(e), produced the best visual results when compared to the ground-truth in Figure~\ref{fig:results}(f).
\subsection{Quantitative Comparisons}

We select some popular image quality metrics including $\boldsymbol\ell_{1}$, $\boldsymbol\ell_{2}$, Peak Signal to Noise Ratio (PSNR), SSIM to evaluate the performance quantitatively. Table \ref{table:result1} shows the results from our experiment compared to state of the art \cite{pathak2016context, liu2018image} for image inpainting with our S-WGAN in bold.
\setlength{\tabcolsep}{4pt}
\begin{table}[h]
	\vspace{-0.2cm}
		\caption{Quantitative comparison of various performance assessment metrics on 3,000 test images from the CelebA-HQ  dataset. $\dagger$ Lower is better.  $\uplus$ Higher is better.} 
	\label{table:result1}	\centering
	\begin{tabular}{|c|c|c|c|c|}
		\hline
		Method   & $\boldsymbol\ell_{2}$ $\dagger$	& $\boldsymbol\ell_{1}$ $\dagger$ & PSNR $\uplus$ & SSIM $\uplus$\\	
		\hline	
		\textbf{WGAN}  & 3562.13 & 87.03 & 13.50&0.56 \\
		\hline
		 \textbf{CE} & 133.481 & 129.30&27.71 &0.76 \\
		\hline
		\textbf{PConv} & 124.62 & 105.94 &28.82 &0.90 \\  
		\hline 
		\textbf{S-WGAN}  & \textbf{81.03} & \textbf{66.09} & \textbf{29.87} &\textbf{0.94}\\
		\hline
	\end{tabular}
\end{table}

For $\boldsymbol\ell_{2}$ and $\boldsymbol\ell_{1}$, the lower the value, the better the image quality. $\boldsymbol\ell_{2}$ measures the average squared intensity difference of pixels while $\boldsymbol\ell_{1}$ measures the magnitude of error between the ground-truth image and the reconstructed image. Conversely, for PSNR and SSIM, the higher the value, the closer the image quality to the ground-truth. Based on observation from Table~\ref{table:result1}, S-WGAN achieves lower $\boldsymbol\ell_{1}$, $\boldsymbol\ell_{2}$, higher PSNR and higher SSIM values in comparison with \textbf{CE} \cite{pathak2016context} and \textbf{PConv} \cite{liu2018image}, which suggests that S-WGAN provide more accurate predictions than the state-of-the-art inpainting algorithm. 
\begin{figure*}
	\vspace{-0.2cm}
	\centering 
	\subfloat[]{\includegraphics[width=0.168\textwidth]{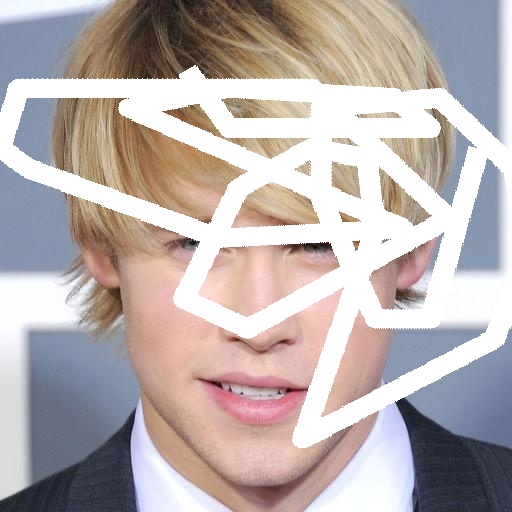}} 
	\subfloat[]{\includegraphics[width=0.168\textwidth]{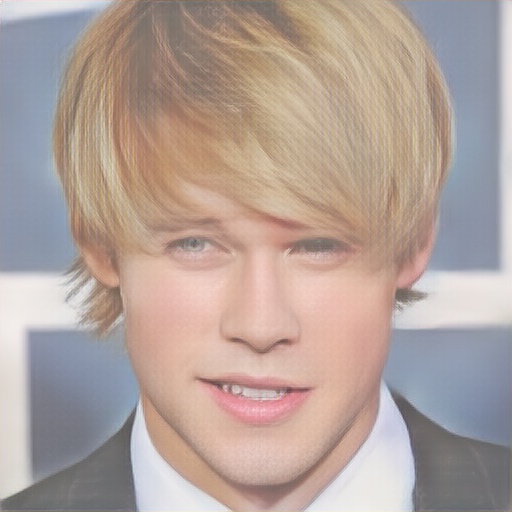}}
	\subfloat[]{\includegraphics[width=0.168\textwidth]{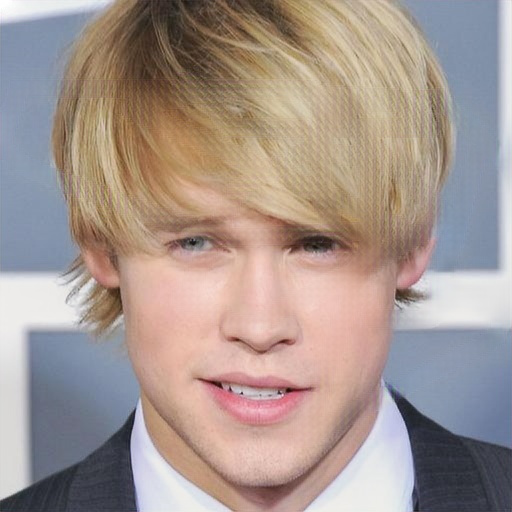}} 
	\subfloat[]{\includegraphics[width=0.168\textwidth]{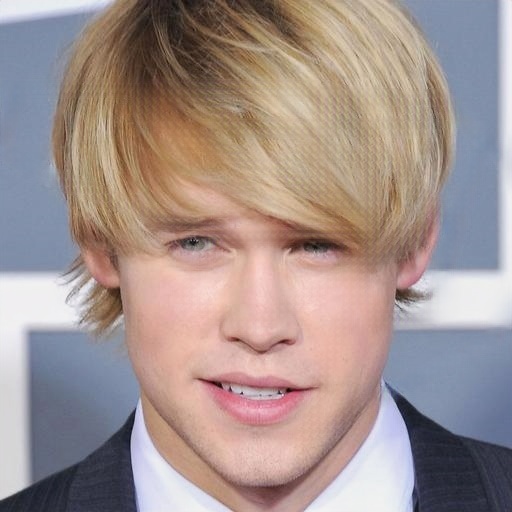}}
	\subfloat[]{\includegraphics[width=0.168\textwidth]{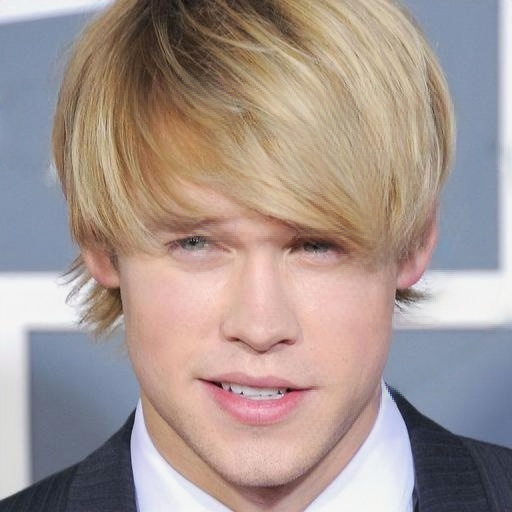}}
	\subfloat[]{\includegraphics[width=0.168\textwidth]{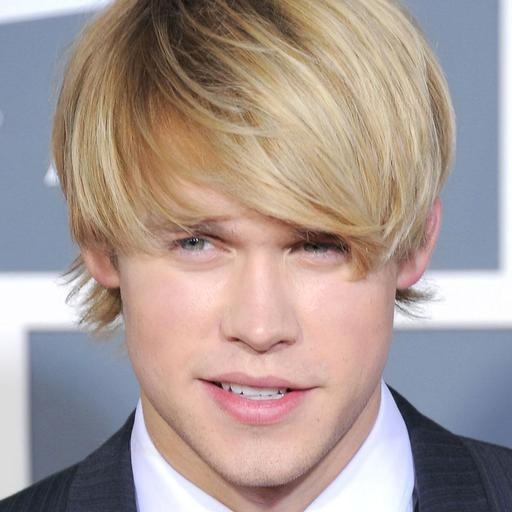}}\\[-5ex]
	\clearsubcaptcounter
	\subfloat[\textbf{INPUT}]{\includegraphics[width=0.168\textwidth]{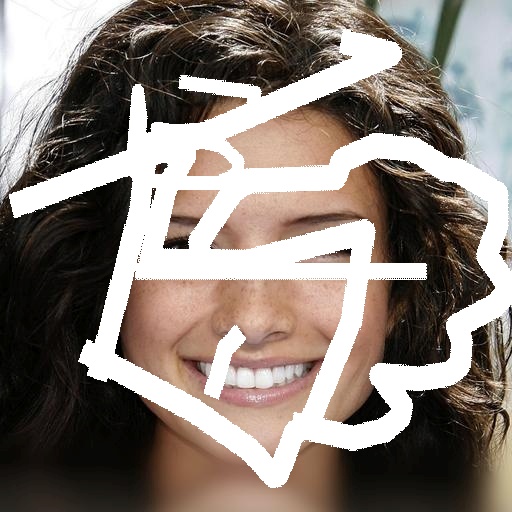}} 
	\subfloat[\textbf{WGAN}]{\includegraphics[width=0.168\textwidth]{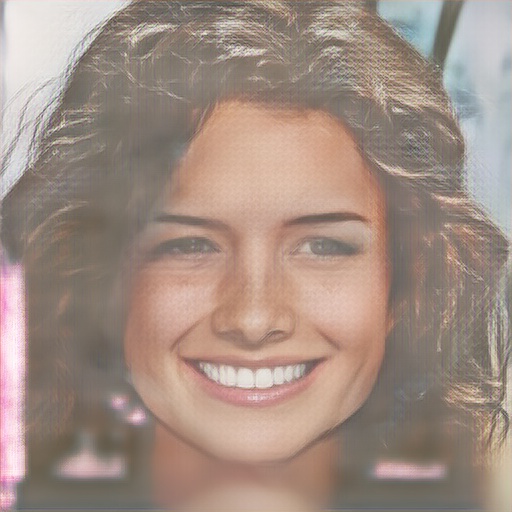}}
	\subfloat[\textbf{WGAN-S}]{\includegraphics[width=0.168\textwidth]{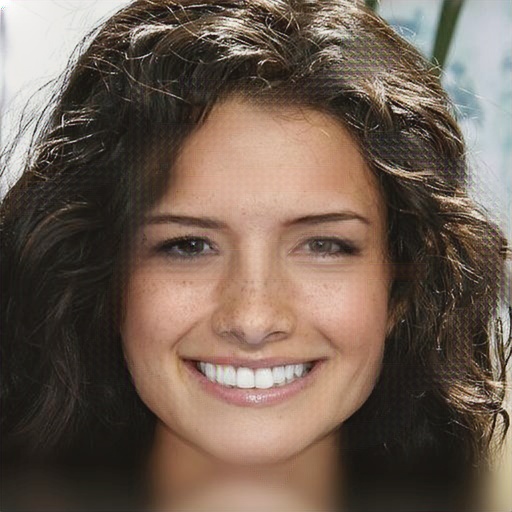}} 
	\subfloat[\textbf{WGANSD}]{\includegraphics[width=0.168\textwidth]{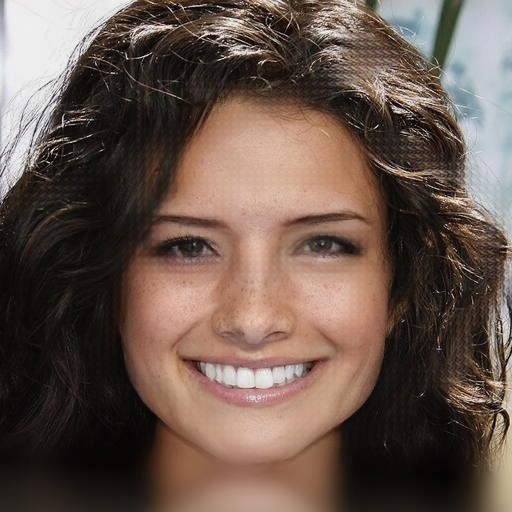}}
	\subfloat[\textbf{S-WGAN}]{\includegraphics[width=0.168\textwidth]{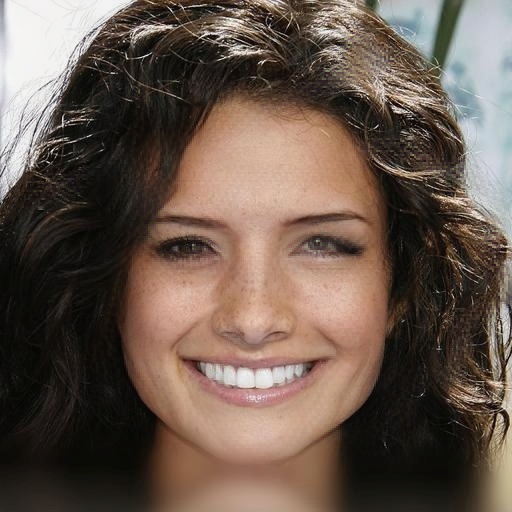}}
	\subfloat[\textbf{GT}]{\includegraphics[width=0.168\textwidth]{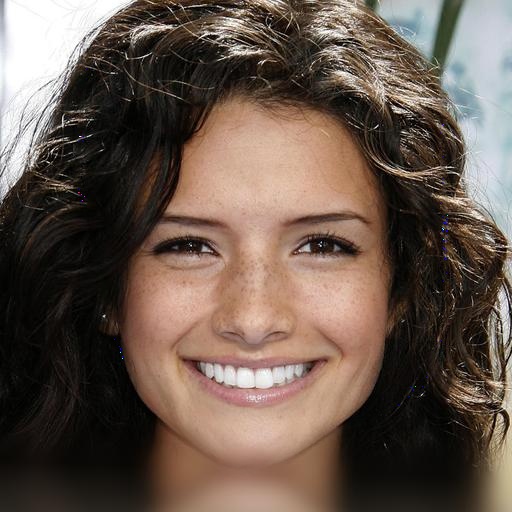}}
	
	\caption{\label{fig:ablations} Qualitative comparison of results using different architectures with the perceptual loss \cite{johnson2016perceptual} on CelebA-HQ \cite{karras2017progressive}. (a) Input masked image; (b) inpainted image by WGAN; (c) Improved WGAN with skip connection; (d) improved WGAN with skip connection and dilated convolution (e) Complete network with $L_{p}$; (f) The ground-Truth image.}
	\vspace{-0.1cm}
\end{figure*}

\section{\uppercase {Ablation Study}}
\noindent To justify the S-WGAN framework and validate the effectiveness of $L_{p}$, we conduct experiments and show intermediate results using different alterations of the S-WGAN on CelebA-HQ dataset. 

Firstly, we conduct investigations on the WGAN and WGAN with skip connection (WGAN-S) using the $L_{f}$, and observed a slight improvement in texture and structure of the reconstructed masked regions of the images. Figure~\ref{fig:ablation} (b) and (c) show changes influenced by skip connections. We observed that visually and quantitatively, the WGAN-S performs better than WGAN model but not satisfactory as shown in the first part of Table~\ref{table:result2}. 

Secondly, we improve the WGAN-S model by including dilated convolutions to each block, and additional convolution layers to obtain our WGANSD model. 
We train the WGANSD with the $L_{f}$ and train the S-WGAN model with our new combined loss function. We noticed that training with the $L_{f}$ improved our results slightly, but not satisfactorily. To verify the differences of these models, we conduct a qualitative and quantitative evaluation. Visually, within the yellow rectangle on Figure~\ref{fig:ablation} comparing columns (d) and (e), the S-WGAN result in column (e) improved with significantly enhanced local detail when compared with column (d) and the original on column (f). Also, in quantitative evaluation shown in Table~\ref{table:result2}, we observe that S-WGAN trained end-to-end with  $L_{wp}$ predicts reasonable outputs with finer details. We also show more qualitatively results in Figure~\ref{fig:ablations} to demonstrate the S-WGAN produces images with preserved realism.
\setlength{\tabcolsep}{4pt}
\begin{table}[h]
	\vspace{-0.2cm}
	\caption{Quantitative difference of results based on different architectures (WGAN), WGAN-S,  WGANSD with $L_{f}$, and S-WGAN trained with $L_{p}$. $\dagger$ Lower is better. $\uplus$ Higher is better.}
	\label{table:result2}\centering
	\begin{tabular}{|c|c|c|c|c|}
		\hline
		 Method   & $\boldsymbol\ell_{2}$ $\dagger$	& $\boldsymbol\ell_{1}$ $\dagger$ & PSNR $\uplus$ & SSIM $\uplus$\\	
		\hline	
		WGAN & 3562.13 & 87.03 & 13.50&0.56 \\
		\hline
		WGAN-S  & 151.4 & 69.59& 27.01&0.87 \\
		\hline
		WGAN-SD  & 145.82 &65.15  &29.26 &0.92 \\
		\hline 
		S-WGAN  & \textbf{81.03} & \textbf{66.09} & \textbf{29.87} &\textbf{0.94}\\
		\hline
	\end{tabular}
\end{table}
\begin{figure*}
	\vspace{-2.0cm}
	\centering
	\subfloat[\textbf{INPUT}]{\includegraphics[width=0.168\textwidth]{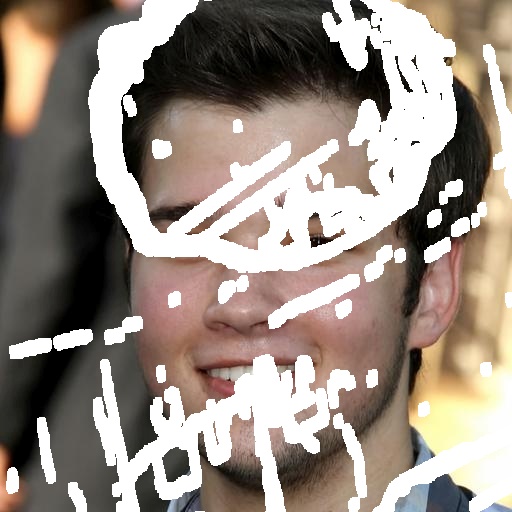}} 
	\subfloat[\textbf{WGAN}]{\includegraphics[width=0.168\textwidth]{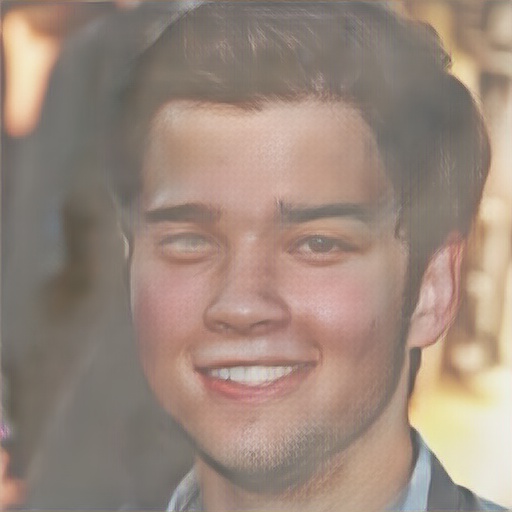}}
	\subfloat[\textbf{WGAN-S}]{\includegraphics[width=0.168\textwidth]{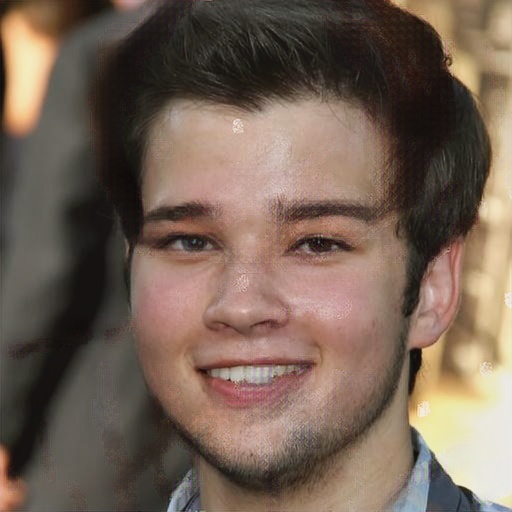}} 
	\subfloat[\textbf{WGANSD}]{\includegraphics[width=0.168\textwidth]{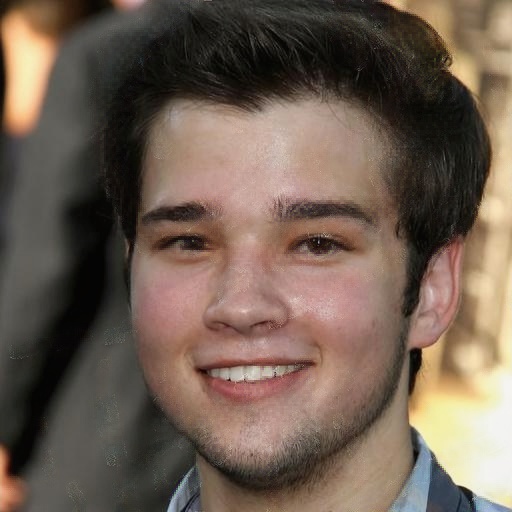}}
	\subfloat[\textbf{S-WGAN}]{\includegraphics[width=0.168\textwidth]{1664swgan.jpg}}
	\subfloat[\textbf{GT}]{\includegraphics[width=0.168\textwidth]{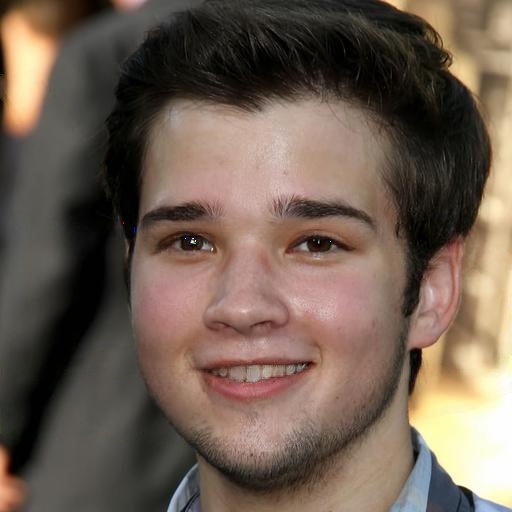}}
	\\[-5ex]
	\clearsubcaptcounter
	\subfloat[\textbf{INPUT}]{\includegraphics[width=0.168\textwidth]{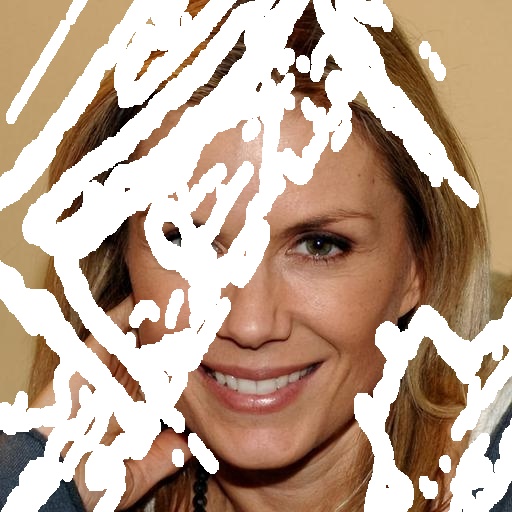}} 
	\subfloat[\textbf{WGAN}]{\includegraphics[width=0.168\textwidth]{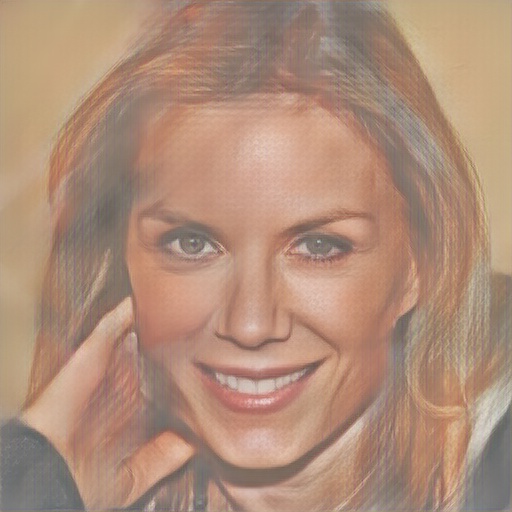}}
	\subfloat[\textbf{WGAN-S}]{\includegraphics[width=0.168\textwidth]{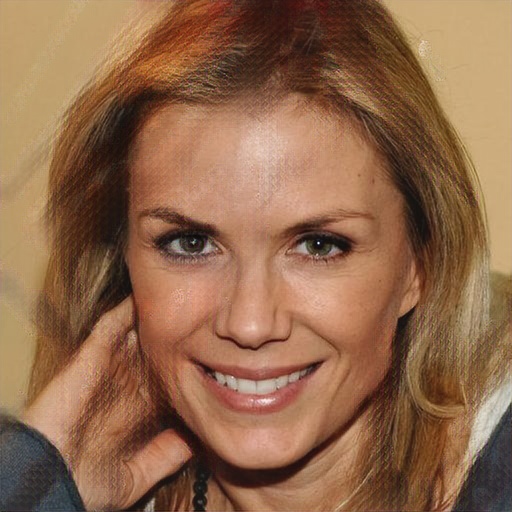}} 
	\subfloat[\textbf{WGANSD}]{\includegraphics[width=0.168\textwidth]{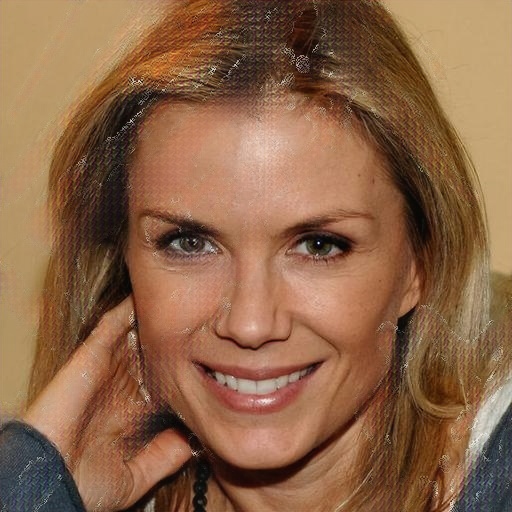}}
	\subfloat[\textbf{S-WGAN}]{\includegraphics[width=0.168\textwidth]{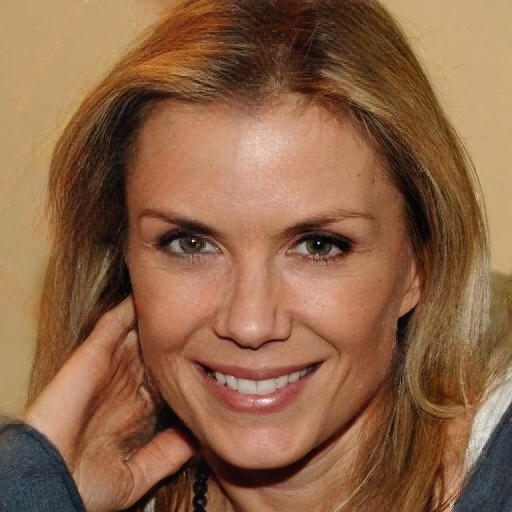}}
	\subfloat[\textbf{GT}]{\includegraphics[width=0.168\textwidth]{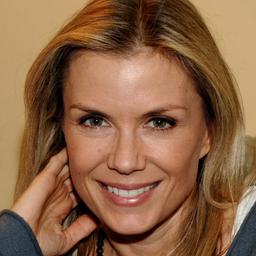}}
	\caption{\label{fig:nvidiamask} Qualitative evaluation of different architectures with perceptual loss \cite{johnson2016perceptual} on CelebA-HQ \cite{karras2017progressive} and Nvidia Mask. (a) Input masked image; (b) Inpainted image by WGAN; (c) Improved WGAN with skip connection; (d) Improved WGAN with skip connection and dilated convolution; (e) Complete network with $L_{p}$; (f) The ground-Truth image.}
	\vspace{-0.1cm}
\end{figure*}
To validate our S-WGANs' representational ability generalised to other masks e.g. Nvidia mask \cite{karras2017progressive}, we use the various architectures of our model to conduct experiments during the ablation studies. We apply the Nvidia mask as the masking method and show our results in Figure~\ref{fig:nvidiamask}.
\section{\uppercase {Discussion}}
\noindent Our proposed S-WGAN with dilated convolution and skip connections trained end-to-end with Wasserstein-perceptual loss function outperforms the state-of-the-art. Our model can learn the end-to-end mapping of input images from a large-scale dataset to predict missing pixels of the binary mask regions on the image. Our S-WGAN automatically learns and identifies missing pixels from the input and encodes them as feature representations, to be reconstructed in the decoder. Skip connections help to transfer image details forwardly and find local minimum by backward propagation. 

Our experiments show the benefit of skip connection combined with Wasserstein-perceptual loss for image inpainting. We have visually compared our proposed method with state of the art \cite{pathak2016context,liu2018image} in Figure~\ref{fig:results}. 
To verify the effectiveness of our network, we carried out experiments with regular convolutions and used the  $L_{f}$. We noticed that the images produced had checkboard artefacts with pitiable visual similarity compared to the original image, as shown in Figure~\ref{fig:results}(d). We introduced skip connections with dilated convolution and our new loss function and obtained improved results that are were semantically reasonable with preserved realism in all aspects.

Compared to existing methods, the generator of our S-WGAN learns specific structures in natural images by minimising $L_{p}$  with an enhanced hallucinating ability powered by symmetric skip connections. Based on Figure~\ref{fig:results}, our S-WGAN can handle irregularly shaped binary mask without any blurry artefacts and has shown edge-preserving and mask completion at border regions on the output images. Additionally, using the Wasserstein discriminator enables the overall network to perform better. This boost the experimental performance of our network to achieve state-of-the-art results in inpainting task on high-resolution images.

One limitation is a consistent practice of other inpainting methods in the preprocessing step. Most preprocessing ignores the fact that the image has to be converted into normalised floating points representations and an inverse-normalisation on the output image, which contributes to the colour discrepancies on the output image, that leads to expensive post-processing. We have been able to solve this using S-WGAN with a new combination of the loss function that preserves colour and image detail. 
\section{\uppercase{CONCLUSION AND FUTURE WORK}}
\label{sec:conclusion}
\noindent In this paper, we propose S-WGAN. Our network can generate images, which are semantically and visually plausible with preserved realism of facial features. We achieved this with a network structure that can widen the receptive field in each block to capture more information and forward to the corresponding deconvolutional blocks. Additionally, we introduced a new combined loss function based on luminance and feature space combined with Wasserstein loss. Our network was able to generate high-resolution images from input covered with arbitrary binary mask shape and achieve a better performance compared to the state-of-the-art methods. The proposed network has shown the effectiveness of skip connections with dilated convolutions as a capture and refining mechanism of contextual information combined with WGAN. For future work, we aim to extend our model to inpaint coarse and fine wrinkles extracted from wrinkle detectors \cite{yap2018automated} with preserved realism.
%
\section*{\uppercase{Acknowledgements}}

\noindent \thanks{We gratefully acknowledge the support of NVIDIA Corporation with the donation of the Quadro P6000 used for this research.}

\bibliographystyle{apalike}
{\small
\bibliography{visappbib}}

%
%

\end{document}